\documentclass[journal,onecolumn]{IEEEtran}
\usepackage[utf8]{inputenc}
\usepackage[T1]{fontenc}
\usepackage[hmargin=25mm,vmargin=27mm]{geometry}
\usepackage{tikz,hyperref}
\usepackage{subfigure}
\ifCLASSINFOpdf
\else
\fi
\begin{document}
\title{Review of Pedestrian Trajectory Prediction Methods: Comparing Deep Learning and Knowledge-based Approaches
} 

\author{Raphael~Korbmacher
        and Antoine~Tordeux
\thanks{R. Korbmacher and A. Tordeux are with the Chair of Traffic Safety and Reliability, University of
Wuppertal, 42119 Wuppertal, Germany (e-mail: \href{mailto:korbmacher@uni-wuppertal.de}{korbmacher@uni-wuppertal.de},  \href{mailto:tordeux@uni-wuppertal.de}{tordeux@uni-wuppertal.de}).}}
\maketitle

\begin{abstract}
In crowd scenarios, predicting trajectories of pedestrians is a complex and challenging task depending on many external factors. The topology of the scene and the interactions between the pedestrians are just some of them. Due to advancements in data-science and data collection technologies deep learning methods have recently become a research hotspot in numerous domains. Therefore, it is not surprising that more and more researchers apply these methods to predict trajectories of pedestrians. This paper compares these relatively new deep learning algorithms with classical knowledge-based models that are widely used to simulate pedestrian dynamics. It provides a comprehensive literature review of both approaches, explores technical and application oriented differences, and addresses open questions as well as future development directions. Our investigations point out that the pertinence of knowledge-based models to predict local trajectories is nowadays questionable because of the high accuracy of the deep learning algorithms. Nevertheless, the ability of deep-learning algorithms for large-scale simulation and the description of collective dynamics remains to be demonstrated. Furthermore, the comparison shows that the combination of both approaches (the hybrid approach) seems to be promising to overcome disadvantages like the missing explainability of the deep learning approach.

\end{abstract}

\IEEEpeerreviewmaketitle

\section{Introduction}
\label{sec:1}
The prediction of future trajectories of pedestrians is a valuable but challenging task. Traditionally, researchers try to model and understand human behavior with simple rules and mechanisms that can be used to simulate realistic behavior and predict future trajectories \cite{Helbing1995, Burstedde2001a}. In the last decade, deep learning methods, that have the interesting property of being able to learn complex features from data alone, have caught a lot of attention in a variety of domains \cite{ourmazd2020science,saporta2008models}. They have led to a large number of practical applications, especially in machine perception, and have shown to outperform the prediction accuracy of the traditional models in many scientific disciplines. %
Prominent examples occur in theoretical biology and active matter \cite{senior2020improved,Cichos2020}, medicine \cite{esteva2017dermatologist,gulshan2016development}, or 
 materials and chemical science \cite{schutt2017quantum,gilmer2017neural,butler2018machine,reyes2019machine}.

In the discipline of pedestrian dynamics, there is a high interest in making accurate predictions of pedestrian trajectories due to numerous real-world applications like facility, infrastructure, and building design \cite{bitgood2006analysis}, notably in the case of evacuation \cite{Boltes2018, Helbing2002,zheng2009modeling}, autonomous driving cars \cite{poibrenski2020m2p3}, human-robot interactions \cite{scheggi2014cooperative}, assistive technologies in industrial scenarios \cite{leo2017computer}, and entertainment (e.g., augmented and virtual reality) \cite{rodin2021predicting}.

\begin{figure}[!ht]
 \centering
 \subfigure[]{\includegraphics[width=0.3508\linewidth]{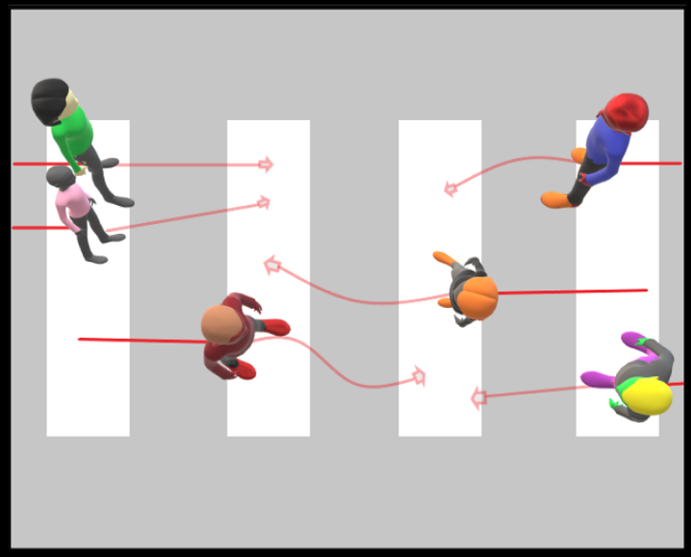}}
 \subfigure[]{\includegraphics[width=0.3592\linewidth]{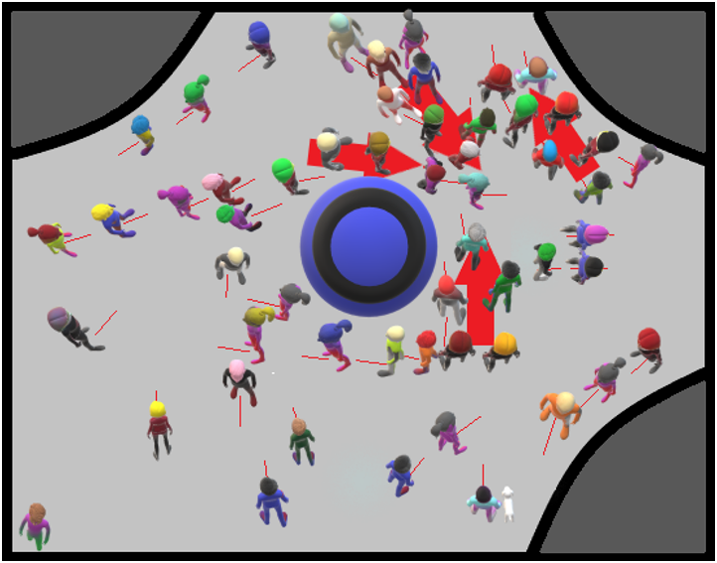}}
 \caption{For the prediction of pedestrian trajectories, it is useful to distinguish between scenes with few pedestrians (see (a)) and crowds of pedestrians (see (b)). Deep learning algorithms turn out to be accurate prediction tools for the scene (a), whereas knowledge-based approaches allow to describe collective phenomena at higher scales such as those described in the scene (b).}
 \label{fig:1} 
\end{figure}
A trajectory is defined as the time-profile of the pedestrian's position. Changes of the position in a given time step can be interpreted as the velocity. Predicting trajectories means to assess the future motion of pedestrians in a given scene. Besides the prediction of the simple physical motion of a single human, based on current or past observations, an important issue is to take into account their possible interactions as well as arising collective dynamics. Collective dynamics can be considered by predicting trajectories of many pedestrians in a scene simultaneously. In Fig.~\ref{fig:1}(a) few pedestrians are present. The challenge consists in predicting individual trajectories in local interactions with the neighbors over relatively small horizon times. In Fig.~\ref{fig:1}(b) a pedestrian crowd is shown. Predicting multiple trajectories and resulting collective crowd behavior at higher scale is necessary.

One approach to tackle the challenge of predicting trajectories is to use the knowledge-based (KB) approach, sometimes referred to by other names, such as physics-based \cite{tordeux2020}, reasoning-based \cite{kruse2013}, expert-based \cite{cheng2020trajectory} or traditional approach \cite{Bighashedel2019}.
The KB approach contains models that are defined by basic rules or generic functions and considers physical as well as social or psychological factors of the pedestrians. They can be specified by a few parameters which generally have physical \emph{interpretations} and allow to adjust the model. 
Prominent microscopic KB models are the social force model (SFM) \cite{Helbing1995} or the optimal reciprocal collision avoidance (ORCA) \cite{vanderBerg2011}.
In general, the goal of these models is to improve the understanding of pedestrian dynamics and to identify microscopic mechanisms and parameters for the emerge of patterns and organisation in given scenes \cite{10.1007/978-1-4419-9725-8_52,Helbing2005,Chowdhury2000,Bellomo2012,Moussa2012,Shahhoseini2017,Boltes2018}. 
These models can be successfully applied for trajectory predictions in both scenes shown in Fig.~\ref{fig:1}. 

The other promising possibility to predict pedestrian trajectory consists of the use of \emph{supervised deep learning} (DL) methods, especially long short-term memory networks (LSTM), convolutional neural networks (CNN), and generative adversarial networks (GAN). In the literature, this approach is often referred to as learning-based \cite{kruse2013}, pattern-based \cite{rudenko2020human}, or data-based approach.
Deep learning refers to methods for training neural networks with more than two hidden layers (deep neural networks) \cite{frochte2019maschinelles} and supervised to methods that learn from data. 
Improving trajectory predictions for autonomous vehicles, service robots, and urban video surveillance are main goals \cite{rudenko2020human}. In opposite to the KB approach, there are no interpretable parameters and rules necessary, but large amounts of data and flexible algorithms.
   
\begin{figure*}[!ht]
  \centering
  \input{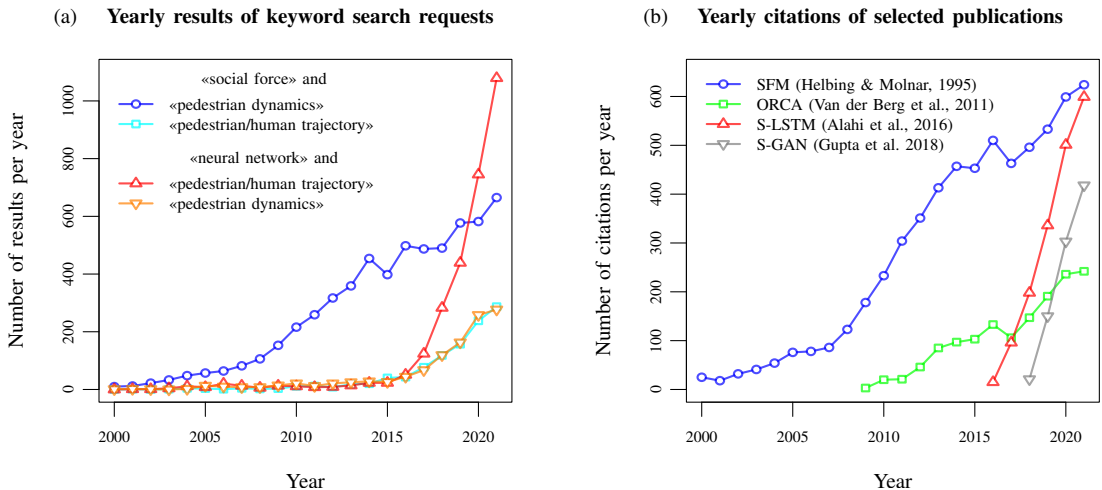}
  \caption{Number of annual citations estimated by performing an online search with the engine Google Scholar \cite{googlescholar}. Panel (a): Results of the tags \guillemotleft neural network\guillemotright\ and \guillemotleft social force\guillemotright\ in combination with either \guillemotleft pedestrian/human trajectory\guillemotright\ or \guillemotleft pedestrian dynamics\guillemotright.
  Panel (b): Yearly citations for the articles by Helbing and Moln\'ar \cite{Helbing1995} (SFM), Van den Berg et al.\ \cite{vanderBerg2011} (ORCA), Alahi et al.\ \cite{Alahi2016} (S-LSTM), and Gupta et al.\ \cite{gupta2018} (S-GAN). SFM and ORCA are knowledge-based, whereas S-LSTM and S-GAN are deep learning approaches.}
  \label{fig:2} 
\end{figure*}

The KB approach is scientifically relevant since several decades, see for illustration in Fig.~\ref{fig:2} the citation numbers of one of the most famous KB models, the social force model by Helbing and Molnár \cite{Helbing1995}. On the other hand, the DL approach is a youthful methodology that started to become highly relevant for the prediction of pedestrian trajectories after the publication of the social-LSTM from Alahi et al. \cite{Alahi2016}. After this date, the social force model started to get associated with pedestrian trajectory as well, see the publication number with the keywords combining social force and pedestrian trajectory in Fig.~\ref{fig:2}.
Yet, the applications of KB and DL approaches are not identical. 
KB models are mainly designed to describe and understand collective phenomena implying high numbers of agents in high density scenarios like in Fig.~\ref{fig:1}(b). They are based on few parameters that can be interpreted and calibrated. 
This results in high flexibility of the prediction scenarios, and notably the possibility to change pedestrian motion preferences (e.g., higher desired time-gap during pandemics). 
DL methods are designed to predict pedestrian trajectories at very local scales in space (few meters) and time (few seconds). 
If their predictions are accurate, they do not allow to control the motion and to reproduce different kinds of behaviors. 
In general, they are used for low density scenarios at local scale like in scene (a) presented Fig.~\ref{fig:1}. 
The question of whether DL approaches could also be suitable for large-scale simulation or successfully used to initiate collective dynamics is still open. 
On the other hand, the pertinence of KB models to predict local trajectories is, regarding the high accuracy of DL algorithms, questionable.

The combination of both KB and DL approaches seems to be especially promising.
Recently, some authors try to implement components of the KB models in the DL algorithms to overcome crucial limitations of the DL approach, like the lack of interpretability or generality \cite{antonucci2020generating, hu2016harnessing,ALAHi2021anchors}. Furthermore, another possibility is to use DL methods to improve the accuracy of the KB models. This can be done by estimating the parameters based on the results of an DL algorithm or to implement a neural network in the KB simulations \cite{zhang2020direction,kielar2020artificial}.
These combinations are called the hybrid approach and they benefit from the strengths of both approaches and avoid their shortcomings.
Other promising algorithms close to hybrid approaches rely on reinforcement learning (RL) and inverse reinforcement learning (IRL). 
The agents learn from their own experiences in RL methods while IRL partly rely on data. 

In this article, we address a comprehensible bibliographical review of KB and DL approaches for the modeling and prediction of pedestrian trajectories. 
We critically compare the two modelling approaches from their technical aspects as well as their application fields. 
We highlight similarities and differences between the two methodologies and draw future development perspectives.
The manuscript is organised as follows. 
A thorough literature review of microscopic KB pedestrian models is presented in the next section. In Sec.~\ref{sec:3}, we give a literature overview about the DL approach distinguishing between long short-term memory networks, convolutional neural networks, and generative adversarial neural networks. 
In Sec.~\ref{sec:4} we show a comprehensive comparison of both approaches focusing on their methodologies, phenomena of interest, and application scales. 
Finally, in the last section, we discuss future directions and perspectives of common developments of the KB and DL modelling approaches. 

\section{The knowledge-based approach}
\label{sec:2}
In the beginnings of the discipline, pedestrian dynamics researchers mostly used direct observations, photographs and time-lapse films to improve knowledge about the behavior of pedestrians \cite{Oeding1963}. This knowledge was applied to develop level-of-service concepts, design elements of pedestrian facilities and planning guidelines \cite{fruin1970designing,polus1983pedestrian,Helbing2005,holl2019level}. These concepts and guidelines are useful to understand and control pedestrian dynamics, but are not suited for predictions of pedestrian flows or trajectories.
Therefore, in the next step researchers started to create simulation models like force-based microscopic models \cite{hirai1975simulation}, queuing models \cite{Lovas1994}, the transition matrix model \cite{garbrecht1973} or the models of Henderson \cite{Henderson1971a,Henderson1974} which conjectured that the behavior of pedestrian crowds is similar to that of gases or fluids. These last models describe aggregated quantities and not individual pedestrian performances.
They are called macroscopic models. 
Currently, KB pedestrian models range from macroscopic, mesoscopic, and microscopic models among others modeling scale characteristics. 
Macroscopic and mesosopic approaches are borrowed from continuous fluid dynamics or gas-kinetic models describing the dynamics at an aggregated level, while microscopic approaches model individual pedestrian motions.
In the literature, many reviews focus on the modelling scales of pedestrian dynamics and the passages from a modelling scale to another \cite{Chowdhury2000,Helbing2001,Castellano2009,bellomo2011modeling,martinez2017modeling,Chraibi2018}.
Some other reviews highlight pedestrian collective dynamics \cite{Schadschneider2009,DUIVES2013193,Boltes2018,Schadschneider2018} or applications in layout design \cite{dong2019state}. 
In the following, we propose a review of KB pedestrian models focusing on microscopic approaches and their applications for prediction of pedestrian trajectories.

\subsection{Microscopic pedestrian models}
In the three last decades, many researchers focused on individual motion of pedestrians using different microscopic models. 
One of the advantages of microscopic approaches compared to macroscopic ones is their natural ability to reproduce heterogeneous behaviors.
Indeed, the pedestrian being individually considered, it is straightforward to attribute specific characteristics to each agent and to take into account behavioral heterogeneity, among other heterogeneous aspects. 
On the other hand, microscopic models can be computationally expensive and their use is limited in case of large-scale simulation.

Microscopic pedestrian models consider the individual behaviors and interactions between individuals. 
The crowd dynamic phenomena result from mutual influences at a macroscopic level \cite{Dong2020}. 
Microscopic models are mainly designed to reproduce characteristics macroscopic features such as fundamental diagrams or collective organisations like band formation \cite{cristiani2014multiscale,hoogendoorn2014continuum}. 
These kinds of models, describing individually pedestrian dynamics, can be used to predict trajectories of pedestrians at any scale. 
The individual pedestrian behavior is described according to certain KB rules ground on physical, social or psychological factors \cite{Chraibi2018}. These rules are formulated in hand-crafted dynamic equations based on Newton's laws of motion. 
Given the input information about the initial status of the pedestrians like position, velocity, and acceleration a forward simulation of these rules can be used to predict the future trajectories. 

Depending on the inputs and outputs of the model, the motion of the pedestrians to a new position can be determined in different ways. If the output of the model is the new velocity or acceleration, which then allows calculating the new position, the model is classified as a velocity- or acceleration-based model, respectively. If the position is directly determined by certain rules and is not based on differential equations, the class of models is called decision-based.

\paragraph{Acceleration-based models}
Acceleration-based models, typically \emph{force-based} models, describe a class of microscopic models where the movement of pedestrians is defined by a superposition of exterior forces. 
One of the first force-based models dates backs to the 1975s and the work by Hirai and Tairu \cite{Hirai1975}.
Nowadays, most of the acceleration-based models are force-based consisting of a relaxation (or anisotropic) term to the desired direction and an interaction term. This last term is generally the sum of repulsion (social force) with the neighbours and obstacles \cite{chraibi2011force,Totzeck2020}. 
The interaction force is the gradient of a potential on the distances and eventually the speed differences with the neighbors. 
This gradient may be exponential as in the social force model \cite{Helbing1995}, algebraic as for the centrifugal and generalised centrifugal force models \cite{YuW2005,Chraibi2010}, or partly linear as in the two-dimensional optimal velocity model \cite{Nakayama2005}. 
The interaction force is weighted by a vision field concept, attributing more importance to the obstacles in front.
Acceleration-based models, being of the second order, require relatively fine discretization scheme and may be subjected to numerical pitfalls \cite{Koester2013}.
Many currently developed acceleration-based pedestrian models are extensions of the social force model (see the review \cite{chen2018social} and references therein). 
Yet, not all acceleration-based models rely on force concepts. 
See for instance the model by Moussaid et al. \cite{moussaid2011simple} based on the concept of desired time gap, the model by Karamouzas et al. \cite{Karamouzas2014} relying on time-to-collision variables, or a recent model by Lu et al. \cite{Lu2020} based on anticipation mechanisms.

\paragraph{Velocity-based models}
Acceleration and force-based models allow describing inertial effects, as well as delays in the dynamics. 
Such mechanisms are questionable for pedestrian dynamics. 
Indeed, in contrast to a driver in a vehicle, inertial effects are minor for a pedestrian and almost no latency takes place in the motion process. 
These assertions, upon other, carry the current of \emph{velocity-based} modeling approaches. 
Velocity-based models have been developed since the 2000s in the literature, later than the acceleration-based models \cite{Chraibi2018}. 
They are partly inspired by robotic. 
Technically, velocity-based models rely on first order-differential equations, whereas acceleration-based models are based on second order equations. 
Velocity-based models are speed functions depending on the position differences with neighbors and obstacles. 
As for acceleration-based models, the speed of the neighbors (or speed difference) may be taken into account as well (making the system of speed equations implicit).
A large class of velocity-based models are based on collision cones and collision avoidance techniques \cite{Paris2007,vanderBerg2008,vanderBerg2011,Pellegrini2009,Guy2009,Guy2010,Kim2014,guo2021vr}. 
The models are formulated as optimisation problems on the ensemble of feasible trajectories devoid of collisions \cite{Maury2008,Maury2011}.
The presence of collisions is generally determined by assuming that the velocity of the neighbors is constant. 
The velocity ensemble leading to collisions describes then cones in space \cite{Paris2007}. 
To avoid unrealistic oscillation effect (ping-pong effects), the models are extended to reciprocal velocity obstacle model (RVO) \cite{vanderBerg2008} or optimal reciprocal collision avoidance (ORCA) \cite{vanderBerg2011} for which avoidance techniques are determined in coordination between the pedestrians.
ORCA models and their extensions are frequently used in computer graphics to reproduce crowd behaviors (see, e.g., \cite{yin2019less,charlton2019fast}).
Other velocity-based models derive from concepts of bearing angle \cite{Ondrej2010}, gradient navigation \cite{Dietrich2014}, or, inspired from vehicular dynamics, time gap variable \cite{Tordeux2016,xu2019generalized}.

\paragraph{Decision-based models and cellular automata}

In \emph{decision-based} or \emph{rule-based} models the pedestrian behavior is not modeled based on differential equations, but on rules or decisions determining the new agent positions, velocities, etc. \cite{Chraibi2018}. 
The time is considered to be discrete for this class of models.
In synchronous approaches, the pedestrians make decisions at time $t+\Delta t$  knowing the state of the system at time $t$. 
The time step $\Delta t$, playing the role of reaction time, has a direct physical meaning and can be used for the calibration of the model.
Cellular automata (CA) are typical decision-based models. 
Not only the time is discrete in CA models, the space and state (i.e., velocity) of the pedestrians are discrete as well. 
The pedestrians evolve on a lattice, that is generally squared or hexagonal. 
A cell can be occupied by a single pedestrian only (exclusion rule).
The size of a cell corresponds to the size of a pedestrian, generally 40~cm $\times$ 40~cm on a squared lattice, i.e., a maximal density of 6.25~ped/m$^2$ \cite{Weidmann1993}. 
The first pedestrian CA models date back to the end of the 1990s \cite{FukuiI99A1,Muramatsu1999,Blue2000a,KluepfelMKWS00A}. 
In floor field CA \cite{Burstedde2001a,Burstedde2002,Kirchner2002b,Schadschneider2002}, the rules and transition probabilities to the neighboring cells result from static and dynamic floor fields. 
The static floor field describe the desired velocity of the pedestrian.  The dynamic floor field models the interactions with the neighbors. 
These interactions are inspired from the process of chemotaxis \cite{Benjacob1997} used by some insects, typically pheromones with ants. 
An important modeling part of the CA approaches consists in solving conflict cases when two pedestrians covet the same cell at the same time. 
A priority rule may be defined, which can be random \cite{Kirchner2002b}.
In \cite{Kirchner2003}, friction probabilities are introduced for which no pedestrian reaches the desired cell in case of conflict. 
Such a mechanism allows notably to explain clogging effects at bottlenecks.
Recent decision-based models are based on cognitive effect \cite{von2020cognitive,von2020concept} or learning process \cite{zhang2020direction}.

\subsection{Trends during the past decades}
  
\begin{figure*}[!ht]
  \centering
  \input{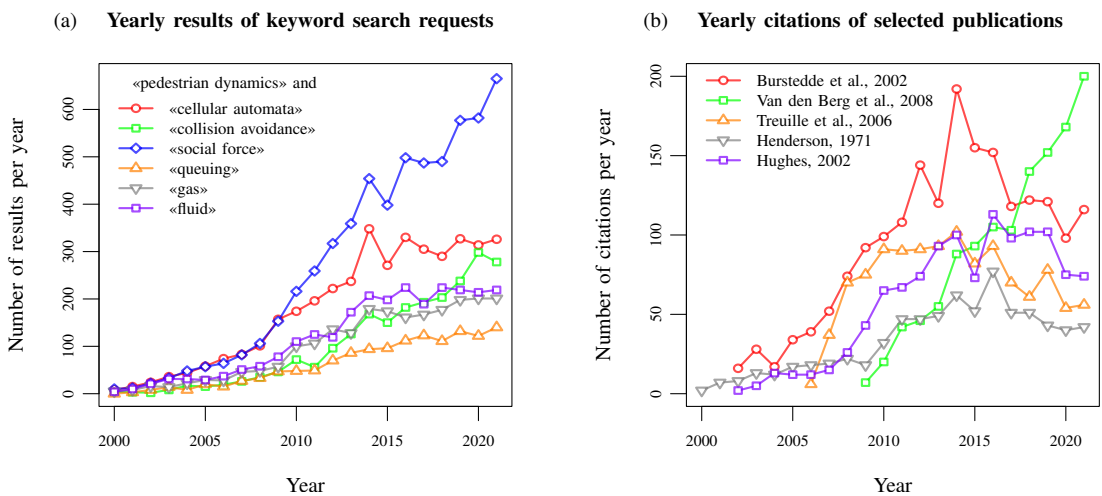}
  \caption{Number of annual citations estimated by performing an online search with the engine Google Scholar \cite{googlescholar}. Panel (a): Results for the tags \guillemotleft pedestrian dynamics\guillemotright \ and keywords related to different model classes. Panel (b): Yearly citations for the articles by  Burstedde et al. \cite{Burstedde2002}, Van den Berg et al. \cite{vanderBerg2008}, Treuille et al. \cite{treuille2006continuum}, Henderson \cite{Henderson1971a}, and Hughes \cite{hughes2002continuum}.}
  \label{fig:3} 
\end{figure*}

The modeling and experimentation of pedestrian dynamics is a quite young research field. First investigations and models date back to the 1960s and the 1970s \cite{Oeding1963,Hirai1975,garbrecht1973,Henderson1971a}.  
However, the topic has mainly been the subject of significant research over the past three decades. 
Experimental studies on pedestrian dynamics in laboratory conditions have been intensively carried out during the 2010s. 
Pedestrian experiments rely on uni-directional flow, counter-flow, bottleneck, intersecting flow, etc. 
An open access data archive can be found in Germany; see \cite{ExpJuelich} and references therein. 
At the same time, authors developed different types of KB pedestrian models, ranging from microscopic to macroscopic modeling scales; see, e.g., the reviews \cite{Chowdhury2000,Helbing2001,Castellano2009,Schadschneider2009,Chraibi2018}. 
Most important in the literature of the KB models is undoubtedly the microscopic social force model by Helbing and Moln{\'a}r, and more generally the force-based microscopic modeling approach (see Table~\ref{tab:1}).  
Traditional KB approaches by cellular automata, queuing processes, or in analogy to fluid or gas dynamics seen currently to reach a plateau, even if the trends are still light increasing (see Fig.~\ref{fig:3}).
The microscopic force-based models and approaches based on collision avoidance techniques remain relevant with an expanding number of citations. 
This is because they commonly serve as benchmark references to evaluate the quality of the predictions with deep learning methods (see, e.g., \cite{Alahi2016,hasan2019}).

\begin{table*}[!ht]
  \caption{Selection of important articles in the literature of knowledge-based pedestrian models with focus on microscopic approaches.\\ \tiny(Number of citations based on search from 31/12/2021)}
  \label{tab:1} 
  \centering
  \begin{tabular}{p{9em}p{34em}p{9em}l}
    \hline
    \textbf{First author, year} & \textbf{Article's title and reference}&\textbf{Family} & \textbf{Citations}\\
   \hline
    Burstedde, 2001&Simulation of pedestrian dynamics using a two-dimensional cellular automaton \cite{Burstedde2001a}&Cellular automata&1931\\
    Kirchner, 2002&Simulation of evacuation processes using a bionics-inspired cellular automaton model for pedestrian dynamics \cite{Kirchner2002b}&---&1119\\
    Van den Berg, 2008&Reciprocal velocity obstacles for real-time multi-agent navigation \cite{vanderBerg2008}&Collision avoidance&1363\\
    Pellegrini, 2009&You'll never walk alone: Modeling social behavior for multi-target tracking \cite{Pellegrini2009}&---&1206\\
    Van den Berg, 2011&Reciprocal $n$-Body Collision Avoidance \cite{vanderBerg2011}&---&1440\\
    Helbing, 1995&Social force model for pedestrian dynamics \cite{Helbing1995}&Force-based&6245\\
    Helbing, 2000&Simulating dynamical features of escape panic \cite{Helbing2000}&---& 5382\\
    Chraibi, 2010&Generalized centrifugal-force model for pedestrian dynamics \cite{Chraibi2010}&---&335\\
    Moussaid, 2011&How simple rules determine pedestrian behavior and crowd disasters \cite{moussaid2011simple}&---&1009\\
    Karamouzas, 2014&A universal power law governing pedestrian interactions \cite{Karamouzas2014}&---&247\\
    Treuille, 2006&Continuum crowds \cite{treuille2006continuum}&Queuing&1176\\
    Henderson, 1971&The statistics of crowd fluids \cite{Henderson1971a}& Gas-kinetic&778\\
    Hughes, 2002&A continuum theory for the flow of pedestrians \cite{hughes2002continuum}& Fluid dynamics&1174\\
    Chowdhury, 2000&Statistical physics of vehicular traffic and some related systems \cite{Chowdhury2000}&Review&2862\\
    Helbing, 2001&Traffic and related self-driven many-particle systems \cite{Helbing2001}&---&3979\\
    Castellano, 2009&Statistical physics of social dynamics \cite{Castellano2009}&---&3963\\
    Schadschneider, 2009&Evacuation dynamics: Empirical results, modeling and applications \cite{Schadschneider2009}&---&766\\
    Bellomo, 2011&On the modeling of traffic and crowds: A survey of models, speculations, and perspectives \cite{bellomo2011modeling}&---&464\\
    Bechinger, 2016&Active particles in complex and crowded environments \cite{bechinger2016active}&---&1507\\
\hline
  \end{tabular}
\end{table*}

\subsection{Knowledge-based models for understanding and predicting}
The aim of knowledge-based models is mainly to identify mechanisms and fundamental parameters operating collectively in the pedestrian dynamics. 
First of all, body exclusion effects are responsible for jamming and clogging effects, and notions of maximal density. 
The parameters of KB models rely on the concept of \emph{fundamental diagram}, a phenomenological uni-modal relationship between the flow and the density.
The fundamental diagram relationship has been pointed out as early as the 1960s.
Yet, investigations on its shape and scattering are still actively ongoing \cite{hankin1958passenger,older1968movement,navin1969pedestrian,seyfried2005fundamental,chattaraj2009comparison,zhang2012ordering,subaih2019gender}. 
Identified parameters are, upon others, the desired speed, the agent size, or the reaction time at microscopic scales. 
They are the maximal density or the capacity at the macroscopic level.
The quantitative estimations of the parameters, as well as their numbers and nature, are controversial and subject to diverse influences, type of the flows (e.g., uni-directional, bi-directional), context and motivation, ages, or cultural effects (see \cite{Boltes2018} and references therein). 
Simple microscopic rules allows explaining macroscopic shapes of the fundamental diagram \cite{schadschneider2009validation,moussaid2011simple,gomes2019parameter}.
Here, as for traffic flow, temporal parameters such as the reaction time or the time gap with the next pedestrian ahead turn out to be highly relevant \cite{moussaid2011simple,Tordeux2016}. 

One of the main highlights of KB models is the identification of self-organisation phenomena and the emergence of coordinated dynamics, patterns, structures, and orders at macroscopic scales.
Multi-scale approaches allow understanding how microscopic individual behaviors initiate the emergence of macroscopic collective dynamics  \cite{cristiani2014multiscale,hoogendoorn2014continuum}. 
Prominent examples of collective dynamics are lane formation 
\cite{Cristin2019,Goldsztein2020}, stop-and-go waves \cite{Bain2019,Friesen2021}, freezing-by-heating effect \cite{Helbing2000a,Stanley00}, herding effect \cite{Helbing2000a,Kirchner2002b} 
or oscillations, intermittent flow, and pattern formation at bottlenecks and in intersections \cite{Helbing2005,Helbing2009,Cividini2013,nicolas2018counterintuitive}; see the review \cite{Chowdhury2000,Bellomo2012,Boltes2018} and references therein.
Comparable self-organisation phenomena arise in social systems and social networks, notably for opinion formation \cite{Hermann2012,Moussaid2013,Touboul2019}. 
This includes a large class of non-equilibrium systems of self-driven or active particles often called \emph{active matter} in the literature of statistical physics \cite{schweitzer2003brownian,Castellano2009,Ramaswamy2010,VICSEK2012,marchetti2013hydrodynamics,Elgeti2015,bechinger2016active}. 
Understanding complex non-linear dynamics operating at different modeling scales remains challenging and currently attracts much attention, notably through data-based approaches \cite{Shahhoseini2017,Needleman2017,Cichos2020,ourmazd2020science,Dulaney2021}. 

The aims of microscopic KB models are mainly to provide a better understanding of large-scale dynamics from individual walking behaviors. 
Predictions of trajectories for a given dataset is no direct goal of KB models, but an indirect one. 
In the literature of pedestrian dynamics, some KB models have been proven to be useful for predicting pedestrian trajectories. 
They are implemented in multi-agent simulation tools  \cite{Korhonen2010,Curtis2016,Taillandier2019,Kleinmeier2019} to analyse pedestrian dynamics in different types of infrastructures or for specific outdoor events. 
The KB model, typically the social force model, is the technical kernel of the dynamics in complement to furthers mechanisms and controls describing different types of behaviors or motivation levels, agent characteristics,  and further context effects. 

\section{The deep learning approach}
\label{sec:3}
The foregoing approach heavily relies on a theoretical modeling framework. Pre-identified key mechanisms were formulated in equations with a few meaningful parameters that have to be calibrated and validated. After that, the model can be used to simulate the scene, which can generate useful information for many applications. 
The predictive capacity for pedestrian trajectories almost comes as a by-product in the KB approach.
In the DL approach the prediction of trajectories is not a by-product, but the main focus.
As in many other domains, the DL approach captured a lot of attention over the last decade due to an increase of real and experimental databases, improvements in the computational capacity of computers \cite{Chraibi2018}, and the requirement of accurate pedestrian predictions for applications like autonomous vehicles or service robots \cite{Rudenko2020}.

There exist different possibilities in the literature to classify the DL methods. Rudenko et al. \cite{Rudenko2020} classify the DL approach into sequential methods and non-sequential methods, based on the type of function approximation they are used for. 
Bighashdel and Dubbelman \cite{Bighashedel2019} classify the methods according to their main focus: The interaction-based approach, where the interactions between the pedestrians are addressed, the path-planning approach, where the trajectories are highly affected by the destinations, and the intention-based approach, where the intention is estimated. 
In this article, we differentiate between three classes of supervised DL algorithms: LSTM, CNN, and GAN. Currently, these three classes of methods seem to be most relevant in research, although there are also promising publications on transformer networks \cite{giuliari2021transformer,isattentionall} or variational autoencoder \cite{ivanovic2020multimodal}.\\
Before researchers start to use these DL algorithms, statistical models were applied to make predictions based on data. These statistical models learn pedestrian behavior by fitting different function estimates to data.
One possibility is to use linear models with Kalman filters or extended Kalman filters \cite{schneider2013pedestrian,meuter2008unscented,Alahi2016}. Kalman filters can be used for predictions by propagating the current state with a dynamical model without the inclusion of new measurements. 
These simple models are not able to account for interactions between humans and thus do not fit for predictions of crowded scenes. The first statistical models that could learn interactions are those based on Gaussian process (GP) like the IGP from Trautman et al. \cite{trautman2015robot}. They proposed an interacting GP where the trajectory of a pedestrian is represented as a Gaussian process. The interaction potential combines multiple trajectories, so that multi-modal distributions can be represented with relatively few parameters.  It has been demonstrated, that they perform well with noisy observations and have closed-form predictive uncertainty. Also in \cite{bennewitz2005learning,Trautman2010, keller2013will,Ellis,kim2011gaussian} GP based models where proposed to model future pedestrian behavior.
Other common statistical approaches that are not based on deep learning are approaches based on Markov property \cite{ferguson2015real}. These approaches include hidden Markov models, in which the hidden state is the pedestrians intent \cite{bennewitz2005learning,kim2011gaussian}. Whereas the GP uses the entire observed trajectory for the prediction of future trajectories, the predictions with Markov models only depend on the current state \cite{ferguson2015real}.\\
Besides these statistical models, reinforcement learning (RL) is an important method for modelling crowd behavior. Most of all, RL techniques are relevant for robotics to anticipate surrounding pedestrian behavior and to plan a collision-free paths \cite{chen2017socially, wan2018robot}. RL is an unsupervised machine learning method, where an objective is learned via trial and error associated to a \emph{reward function} that rewards or penalizes agent behaviors. Therefore, no data is required. It is assumed that the reward function contains the necessary information \cite{martinez2020using}. An exception to this is \emph{inverse} RL (IRL) where the design of the reward function is based on data. Kretzschmar et al. \cite{kretzschmar2016socially} introduce an IRL algorithm that uses a maximum entropy probability distribution for a joint set of continuous state-space for mobile robot navigation in crowds. Kitani et al. \cite{kitani2012activity} propose an well-established IRL algorithm for plausible path predictions. To extend the flexibility of the IRL algorithms recent works use deep IRL algorithms that can estimate non-linear, continuous reward functions. These deep IRL show promising results especially in robot path planning and vehicle driving behavior \cite{wulfmeier2016watch}. If it is the aim to predict pedestrian trajectories, these deep IRL algorithms are often combined with LSTM networks, which are supervised learning algorithms that will be discussed in the following. For more literature about RL in pedestrian dynamics refer to, e.g., \cite{le2016reinforcement, martinez2014marl,martinez2011multi}. For IRL methods used for trajectory predictions see \cite{kitani2012activity,rhinehart2018first}, for deep IRL methods see \cite{lee2017desire, lee2018crowd,everett2021collision}, or to get a wider perspective, refer to the surveys \cite{van2021algorithms, martinez2017modeling}.

\begin{table*}[!ht]
  \caption{Selection of important articles of deep learning algorithms predicting pedestrian trajectories.\\ \tiny(Number of citation based on search from 31/12/2021)}
  \label{tab:2} 
  \centering
  \begin{tabular}{p{9em}p{38em}p{5em}l}
    \hline
    \textbf{First author, year} & \textbf{Article's title and reference}&\textbf{Family} & \textbf{Citations}\\
    \hline
    Trautman, 2010&Unfreezing the robot: Navigation in dense, interacting crowds \cite{Trautman2010}&Gaussian&459\\
    Kitani, 2012&Activity forecasting \cite{kitani2012activity}&IRL&727\\
    Alahi, 2016&Social-LSTM: Human trajectory prediction in crowded spaces \cite{Alahi2016}&LSTM&1770\\ 
    Kretzschmar, 2016&Socially compliant mobile robot navigation via IRL\cite{kretzschmar2016socially} \cite{kitani2012activity}&IRL&326\\
    Yi, 2016&Pedestrian behavior understanding and prediction with deep neural networks \cite{yi2016pedestrian}&CNN&124\\
    Lee, 2017 &DESIRE: Distant future prediction in dynamic scenes with interacting agents \cite{lee2017desire}&LSTM&604\\
    Fernando, 2018&Soft+ hardwired attention: A LSTM framework for human trajectory prediction and abnormal event detection \cite{fernando2018soft}&LSTM&219\\
    Gupta, 2018&Social GAN: Socially acceptable trajectories with generative adversarial networks \cite{gupta2018}&GAN&910\\
    Nikhil, 2018&Convolutional neural network for trajectory prediction \cite{nikhil2018convolutional}&CNN&83\\ 
    Vemula, 2018&Social attention: Modeling attention in human crowds \cite{vemula2018social}&LSTM&357\\
    Xue, 2018&SS-LSTM: A hierarchical LSTM model for pedestrian trajectory prediction \cite{xue2018ss}&LSTM&217\\   
    Sadeghian, 2019&SoPhie: An attentive GAN for predicting paths compliant to social and physical constraints \cite{sophie}&GAN&457\\
    Rudenko, 2020&Human motion trajectory prediction: A survey \cite{Rudenko2020}&Review&282\\
    \hline
  \end{tabular}
\end{table*}

\begin{figure*}[!ht]
  \centering
  \input{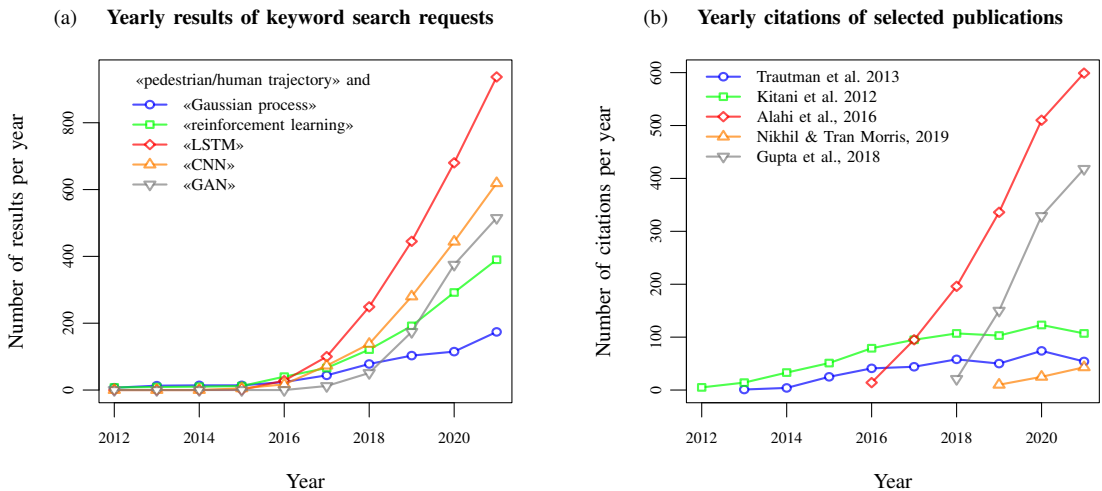}
  \caption{Number of annual citations estimated by performing an online search with the engine Google Scholar \cite{googlescholar}. Panel (a): Results for the tags \guillemotleft pedestrian trajectory\guillemotright\ and keywords related to the class of algorithm. Panel (b): Yearly citations for the articles by Trautman et al. \cite{trautman2015robot}, Kitani et al. \cite{kitani2012activity}, Alahi et al. \cite{Alahi2016}, Yi et al. \cite{yi2016pedestrian}, and Gupta et al. \cite{gupta2018}.}
  \label{fig:4} \vspace{-5mm}
\end{figure*}

\subsection{Long short-term memory networks}
A recurrent neural network (RNN) is an important class of machine learning methods, that uses feedback connections to store representations of recent input events in form of activations \cite{hochreiter1997long}. These feedback connections can exist between different time steps providing a temporal memory to the network \cite{Li2018}. Because of this capability, they are especially suited for sequence modeling tasks such as time series prediction and sequence labeling tasks \cite{Azzouni2017}. Most successful are the LSTM architectures of RNNs, that use purpose-built memory cells to store information. They have achieved impressive results in many sequential prediction tasks like speech recognition \cite{graves2013speech}, machine translation \cite{sutskever2014sequence}, handwriting recognition and generation \cite{graves2013generating}, and image captioning \cite{vinyals2015show}. LSTM networks can be trained for sequence generation by processing real data sequences one step at a time and predicting what comes next so that temporal coherence can be ensured. The training of a LSTM network can be lengthy and difficult, but the forward-propagation of new data (testing) can happen quite fast. That is why they are a popular choice for real-time tracking and predicting.

Recently, many researchers use these LSTM networks to predict trajectories of pedestrians. Most famous is the social-LSTM proposed by Alahi et al. \cite{Alahi2016}. It is the first major DL application that takes the interactions between pedestrians in crowded scenarios into account. The main contribution of the social-LSTM is a novel pooling layer called social pooling that gathers the hidden states of nearby pedestrians. With that technique, the influence of neighboring pedestrians on the movement of the ego pedestrian can be included as an input for the following prediction step.
Inspired by the success of the social-LSTM many researchers try to use LSTM networks with different settings. The scene-LSTM incorporates additional scene information \cite{manh2018scene}. The social-scene LSTM \cite{xue2018ss} considers the scene influence, the social information of the person, and scene scale information in three different LSTM networks. Pfeiffer et al. \cite{Pfeiffer2018} incorporate static obstacles that have to be avoided. Other approaches that process scene information are \cite{lee2017desire, lisotto2019social}.
Besides the capturing of scene information, some LSTM algorithms integrate an attention mechanism in the interaction module to capture the relative importance of each person in the scene.
Other LSTM networks focus on attention mechanisms in the interaction module to capture the relative importance of each person in the scene. In some of these works, the attention weights are learned by data or by added handcrafted based on domain knowledge \cite{fernando2018soft,haddad2019situation,huang2019stgat,tao2020dynamic}. 
Another relevant work is STGAT by Huang et al. \cite{huang2019stgat} that uses graph neural networks (GNN) instead of a pooling module. To share information across the pedestrians, each agent is treated as a node of a graph. Derived from this work DAG-Net \cite{9412114} presents a double attentive GNN combined with a RNN and STR-GGRNN \cite{9506209} introduces an online framework that automatically infers the social interactions by completing the graph edges.
An overview of DL pedestrian trajectory prediction algorithms relying on LSTM networks is provided in Table~\ref{tab:3}. 
The works are sorted by the publication year to give the reader an impression about the overall progress in this field.

\begin{table*}[htbp]
  \centering  
    \caption{Overview of DL pedestrian trajectory prediction algorithms relying on LSTM networks.}
  \label{tab:3}
    \begin{tabular}{p{10em}p{12em}p{32em}}
        \hline
    \textbf{First author, year} & \textbf{Name} & \textbf{Main characteristics} \\
    \hline
    Alahi, 2016 \cite{Alahi2016}& Social-LSTM&Social pooling layer to handle interactions\\
    Lee, 2017 \cite{lee2017desire}&DESIRE & Ranks and refines the generated trajectories\\
    Fernando, 2018 \cite{fernando2018soft}& Soft + Hardwired Attention & Utilises “soft attention” as well as “hard-wired” attention\\
    Xue, 2018 \cite{xue2018ss} & SS-LSTM&Three different LSTMs to capture person, social and scene scale information\\
    Manh, 2018 \cite{manh2018scene}& Scene-LSTM&Incorporates scene information\\
    
    Hasan, 2018 \cite{hasan2018mx} & MX-LSTM  &Takes head pose and vision range into account\\
    Vemula, 2018 \cite{vemula2018social}& Social-Attention &Based on social-LSTM, but captures the relative importance of each person when navigating in crowds\\
    Bisagno, 2018 \cite{bisagno2018group} & Group- LSTM&Coherent filtering algorithm to segment groups\\
    Cheng, 2018 \cite{cheng2018pedestrian} & Social-Grid LSTM&Combines social pooling and grid-LSTM methods\\
    Bartoli, 2018 \cite{bartoli2018context} & Context-aware-Social-LSTM  & Interactions with static elements and dynamic agents\\
    Pfeiffer, 2018 \cite{Pfeiffer2018}& Static-LSTM  & Angular pedestrian grid combined with CNN\\
    Ivanovic, 2019 \cite{ivanovic2019trajectron}& The Trajectron  & LSTMs combined with CVAEs and dynamic spatiotemporal grahical strucutres\\
    Lisotto, 2019 \cite{lisotto2019social}& SNS-LSTM& Social, navigation and semantic pooling mechanism\\  Huang, 2019 \cite{huang2019stgat} & STGAT  &Spatial-Temporal Graph neural network combined with LSTM\\          
    Haddad, 2019 \cite{haddad2019situation}& Situation-Aware-LSTM&Spatio- temporal graph that operates on the local and global contexts\\
    Syed, 2019 \cite{syed2019sseg}& SSeg-LSTM&Semantic segmentation to incorporate scene information\\
    Zhao, 2019 \cite{zhao2019multi} & MATF &Encodes past trajectory and scene context into a Multi-Agent Tensor\\
    Monti, 2020 \cite{9412114} & DAG-Net & Double attentive GNN that deals with past interactions and future goals\\
   \hline
    \end{tabular}
\end{table*}

\subsection{Convolutional neural network}
\label{sec:3.2}
Convolutional neural network (CNN) are machine learning methods mostly used for computer vision tasks like image classification \cite{krizhevsky2012}, object detection, object tracking and image segmentation \cite{bengio2009}. 
For studies of pedestrian dynamic these methods are highly important because of their impressive performance in classifying and tracking of objects like pedestrians or vehicles \cite{fabbri2018learning,ristani2018features,yoon2018online}. Another advantage is the reduced computational load outperforming regular neural networks \cite{mohamed2020social,nikhil2018convolutional}.
However, CNN are not widely used to predict pedestrian trajectories, because these are non-sequential methods, which makes it difficult to design the network input and output \cite{yi2016pedestrian}. They are mostly used for trajectory predictions of road vehicles \cite{math9060660}, the prediction of pedestrian behaviors for autonomous vehicles \cite{rehder2018pedestrian,abughalieh2020predicting}, or pose/action recognition \cite{yan2018spatial,cheron2015p}.
The first CNN designed to model and predict pedestrian trajectories is the “Behavior-CNN” from Yi et al. \cite{yi2016pedestrian}. It is a 3-stage deep CNN that encodes the pedestrian behavior into sparse displacement volumes which can be directly used as network input at the first stage.
The CNN from Nikhil et al. \cite{nikhil2018convolutional} utilizes a highly parallelizable convolutional layer to handle temporal dependencies. Trajectory histories are embedded by means of a fully connected layer. Stacked convolutional layers are used to learn temporal dependencies in a consistent manner. The features from the final convolutional layer are passed through a fully-connected layer which  generates all predicted positions simultaneously.

An extension of the CNNs are the graph CNNs which were first introduced by \cite{DBLP:journals/corr/KipfW16}. Mohammed et al. \cite{mohamed2020social} propose the Social-STGCNN that uses graph CNN by modeling the interactions as a spatio-temporal graph, whose edges model the social interactions between the pedestrians. Dan \cite{dan2020spatial} proposed a graph CNN combined with a LSTM network. The graph CNN extract the feature from the pedestrians and the scene for which every pedestrian is regarded as a node and the relationship between each node and its neighbors is obtained by graph embedding. The LSTM encodes the relationship so that the model can predict nodes trajectories. Other approaches combing CNNs with LSTM networks are \cite{Xu_2018_CVPR,jain2020discrete,9000540}. 
An overview of DL pedestrian trajectory prediction algorithms relying on convolutional neural networks is given in Table~\ref{tab:4}.

\begin{table*}[htbp]
  \centering  
    \caption{Overview of DL pedestrian trajectory prediction algorithms relying on convolutional neural networks.}
  \label{tab:4}
    \begin{tabular}{p{10em}p{12em}p{32em}}
        \hline
    \textbf{First author, year} & \textbf{Name} & \textbf{Main characteristics} \\
    \hline
    Yi, 2016 \cite{yi2016pedestrian}& Behavior CNN  &3-stage deep CNN that creates sparse displacement volumes \\
    Varshneya, 2017 \cite{varshneya2017human}& SSCN  & Static spatial context modeled with CNN\\
    Rehder, 2018 \cite{rehder2018pedestrian} & RMDN &CNN for infering destination from images and position. LSTM for  prediction\\
    Nikhil, 2018 \cite{nikhil2018convolutional} & CNN & Highly parallelizable CNN to handle temporal dependencies\\
    Mohamed, 2020 \cite{mohamed2020social}&Social-STGCNN & Models the interactions as a graph using social spatio-temporal graph CNN\\
    Yu, 2020 \cite{yu2020spatio} & TGConv &Transformer based graph convolu-
tion mechanism\\
    Dan, 2020 \cite{dan2020spatial}& Spatial-Temporal Block &Spatial Temporal Graph CNN combined with LSTM\\
    Ridel, 2020 \cite{ridel2020scene} & COVLSTM & 2-D grid combined with CNN and LSTM\\
    Jain, 2020 \cite{jain2020discrete}& DRF-Net & Discrete residual flow network\\
    Zhang, 2021 \cite{zhang2021social}  & Social-IWSTCNN& CNN with spatial and temporal features\\
    Zhao, 2021 \cite{9501325} & STUGCN &CNN with spatio-temporal graph architecture\\
    Zamboni, 2021 \cite{zamboni2021pedestrian} & Conv2D &2D Convolutional models with different network achitectures\\
    \hline
    \end{tabular}
\end{table*}

\subsection{Generative adversarial networks}
\label{sec:3.3}
A problem that is inherent with predictions of trajectories is the multimodal nature of future pedestrian trajectories. GANs have a high potential to cope with this problem, because of their capabilities to generate multimodal samples \cite{STI_GAN}. With the use of GANs, it is possible to predict a distribution of potential future trajectories and not just the best single trajectory \cite{Trajectron}. 
The architecture of a GAN consists of two part: a generator and a discriminator. In the DL approach, a neural network is trained to match the desired data distribution and achieve a low error rate. In a non-intuitive way, the generator of a GAN is trained so that the error rate of the discriminator increases at first. The generator takes the training data as input and creates different output samples. In turn, the discriminator takes the generated samples and the training data and tries to distinguish whether a given sample belongs to the true data distribution or is generated by the generator. Both components are engaged in a competition similar to a two-player min-max game where each one tries to outsmart the other one \cite{gupta2018,math9060660}. From this process, the generator learns to generate data that resemble the true data distribution. Although the results of the GAN-based methods are promising, there are two main difficulties. First, GANs can be hard to train and second, GAN learning often suffers from mode collapse \cite{arjovsky2017,salimans2016}.
Applied for trajectory prediction, most GANs are combined with LSTM networks. In the social GAN from Gupta et al. \cite{gupta2018}  the generator is composed of an LSTM-based encoder, a context pooling module, and an LSTM-based decoder. The discriminator uses LSTMs as well. The pooling module adopted in the social GAN uses a uniform weight for all surrounding pedestrians. Therefore, it can not distinguish the different effects exerted on a target pedestrian by pedestrians at different distances and traveling at different speeds. 
For that reason, many authors added attention mechanisms. Sadegehian et al. \cite{sophie} add an attention module to assign different soft attention distribution weights to the surrounding pedestrians and the static environment. Using an attention module with a physical and social component and a feature extractor module composed of a CNN and several LSTM network encoder, the model SoPhie learns the interaction information of different agents and extracts the most important information from the neighbors. 
Social Ways \cite{socialway} applies info-GAN \cite{infogan}, which introduces latent code to enhance the multi-modality of prediction and computes the attentive social features to generate a more convincing result. It uses discrimination loss for the discriminator, adversarial loss for the generator and information loss for both. Social-BiGAT \cite{kosaraju2019} relies on BiGAN architecture to help reduce the variance of the predicted trajectory distributions and allow for better generalization. In Lv et al. \cite{lv2021improved} the authors propose a GAN combined with transformer networks which generates trajectory distributions to capture the uncertainty of the predictions. 
However, these methods focus on the trajectory prediction in homogeneous environment without considering the types of road users.  Lai  et al. \cite{lai2020} use an attention module, containing two components, in order to alleviate the issues given by the complexity of a scene with many heterogeneous interacting agents \cite{lai2020}.
An overview of DL pedestrian trajectory prediction algorithms relying on generative adversarial neural networks is proposed in Table~\ref{tab:5}.

\begin{table*}[htbp]
    \caption{Overview of DL pedestrian trajectory prediction algorithms relying on generative adversarial neural networks.}
  \label{tab:5}
  \centering  
    \begin{tabular}{p{10em}p{12em}p{32em}}
        \hline
    \textbf{First author, year} & \textbf{Name} & \textbf{Main characteristics} \\
           \hline
          Gupta, 2018 \cite{gupta2018} & Social-GAN & GAN composed of LSTM encoder, context-pooling module and LSTM decoder \\
          Fernando, 2018 \cite{fernando2018gd} & GD-GAN & GAN for pedestrian trajectory predicting and group detection \\
          Sadeghian, 2019 \cite{sophie}& SoPhie  & GAN that uses path history and scence information \\
    Amirian, 2019 \cite{socialway}& Social Ways & Uses discrimination-, adversarial- and information loss\\
    Kosaraju, 2019 \cite{kosaraju2019} & Social-BiGAT & GAN for multimodal trajectory predictions\\
    Lai, 2020 \cite{lai2020} & AEE-GAN  & Info-GAN architecture with recurrent feedback\\
    Huang, 2021 \cite{STI_GAN} & STI-GAN & Combination of graph attention network and GAN\\
    \hline
    \end{tabular}
\end{table*}

\section{Comparing the approaches}
\label{sec:4}
In this Section, we present a comparison of the KB and the supervised DL approaches. First, we focus on technical differences (Sec.~\ref{sec:4.1}) and second, we describe the differences regarding the applications for predicting pedestrian trajectories (Sec.~\ref{sec:4.2}).
 
\subsection{Technically oriented comparison}

\label{sec:4.1}
There are considerable differences between the two approaches, which makes it difficult to create a common framework that both can share for a fair comparison.
Kothari et al. \cite{kothari2021human} define trajectory predictions as "given the past trajectories of all humans in a scene, forecast the future trajectories which conform to the social norms". This definition fits for the DL approach, but not the KB approach where the past trajectories are not used to predict (simulate) the future ones. In KB methods, inputs describe the current state of the system through, e.g., instantaneous relative position \cite{Helbing1995}, relative velocity \cite{Chraibi2010}, time gap \cite{moussaid2011simple}, or collision-related indicators such as bearing angle \cite{Ondrej2010}, collision cone \cite{vanderBerg2008} or time-to-collision \cite{Karamouzas2014}. 
Moreover, the approaches have different criteria in terms of quality of the fit.  A "good" KB model has few interpretable parameters, can simulate realistic pedestrian trajectories, and improves understanding of the phenomena. To evaluate the performance of a KB method, the parameters have to be calibrated and the model needs to be validated. 
The calibration phase is done using knowledge on the values of the parameters (i.e., 1 or 1.5 m/s for the pedestrian desired speed, see \cite{Weidmann1993}) or is formulated using empirical data as an optimisation problem (by least squares or by maximum likelihood, see, e.g.,  \cite{corbetta2015parameter,lovreglio2015calibrating,Tordeux2016b,Bode2019}). 
The validation of KB methods is generally done using three typical ways: by using fundamental diagrams \cite{schadschneider2009validation,feliciani2016improved}, using data \cite{li2015parameter,robin2009specification,ko2013calibrating} and by comparing the resulting phenomena with real-world self-organization phenomena. 
The goal of the DL approach is to predict trajectories that are as close as possible to the real trajectories. Common key figures for the performance of the DL approaches are the average displacement error (ADE) \cite{Pellegrini2009} and the final displacement error (FDE) \cite{Alahi2016}. The first one averages the Euclidean distance between points of the predicted trajectory and the ground truth that have the same temporal distance and the second one measures the distance between the final predicted position and the ground-truth position. When the two approaches are compared in the literature the ADE and FDE are used as reference (see, e.g., \cite{Alahi2016,hasan2019, gupta2018}).

The algorithms of the DL approach have a large number of parameters that have no direct physical meaning and therefore can be referred to as coefficients. These coefficients have to be trained by one part of the data set (training phase). In general, this is done by using an training algorithm like the back-propagation algorithm \cite{nielsen2015neural}. The aim of this training algorithm is to adjust the network settings in a way, that minimizes the given cost function \cite{gottlich2021optimal}. Common cost functions are the mean square error or the  cross entropy. After the training is complete, the algorithms are feed with the rest of the data set to evaluate predictions (testing phase). 
This cross-validation method allows detecting overfitting phenomena, when the algorithm presents low training error but poor performance for the prediction of new data.  
The training is usually computationally expensive and is made offline, whereas the predictions of the trajectories (i.e., the computation), once the training is completed, are quite fast and can be made online (in real-time) \cite{antonucci2020generating}.
We summarize our findings regarding the technical differences of both approaches in Table \ref{tab:6}.  
\begin{table*}[!ht]
      \caption{Technically oriented comparison of the KB and the DL approach for prediction of pedestrian dynamics.}
  \label{tab:6}
  \centering
    \begin{tabular}{p{10em}p{17.5em}p{17.5em}}
    \hline
    & \textbf{Knowledge-based} & \textbf{Deep learning} \\
          \hline
          \emph{Semantics} & Model & Algorithm \\
          & Parameter & Coefficient \\
          & Calibration &Training\\
          &Validation&Testing\\
    \hline
    \emph{Methodology} & Differential equation systems or cellular automata & Neural network, mostly RNN, LSTM, CNN, GAN \\
     \hline
     \emph{Inputs} & System actual state
     (e.g., pedestrian relative positions, velocities, etc. at time $t$) & Past trajectories discretised over the time interval $[t-T,t]$, $T\approx\;$2--4\,s\\
      \hline
     \emph{Outputs} & Future pedestrian positions at time $t+\delta t$, $\delta t$ ranging from 0.01\,s (discretised differential system) to 0.5--1\,s (cellular automata)& Future pedestrian positions discretised over time interval $[t,t+T^\star]$, $T^\star\approx\;$3--5\,s\\
  \hline
    \emph{Modeling concepts} &Fundamental diagram & Learning human-interaction \\
          &Body exclusion, maximal density& Social pooling \\
          &Desired velocity and social force& Attention mechanisms \\
          &Static and dynamic floor field & Group dynamics\\
          & Collision avoidance techniques, spatial and temporal anticipation mechanisms& Human-space interactions\\
          &Vision angle, bearing angle&\\
          &Temporal interaction indicators (time gap, time-to-collision, time-to-interaction)&\\
     \hline
     \emph{Performance evaluation} & Flow-density relationship& Average displacement error (ADE)\\
        &Comparison to experimental data& Final displacement error (FDE)\\
        &Description of collective behaviors and self-organised phenomena & Modified Hausdorff distance (MHD)\\
        \hline
    \end{tabular}
\end{table*}

\subsection{Application oriented comparison}
\label{sec:4.2}

Besides the technical orientated differences, there are also great differences regarding the applications of both approaches. KB approaches are mostly used for crowded situations. Collective dynamics are described at macroscopic scales.
Applications mainly rely on large-scale simulations to analyze, e.g., infrastructure design or evacuation situations. 
In contrast to that, the DL approach focuses more on single pedestrians and their interactions with other pedestrians or the environment locally and in low density situations. Applications are mainly for automatised mobile systems like autonomous vehicles or industrial robots that have to anticipate the future behavior/trajectory of pedestrians to avoid collisions. 
The DL approach surpasses the KB approach in complexity (i.e., number of inputs as well as number of parameters/coefficients). 
This provides a high flexibility and enables the DL algorithms to learn complex interactions and motion patterns when the amount of data is sufficient. This is especially useful in situations where complexity prohibits the explicit programming of a system´s exact physical nature.

\paragraph{The importance of data} 
There are great differences in the role of data for the two KB and DL approaches. The performance of the DL algorithms highly depends on the quality of the data. The majority of researchers in this discipline use datasets containing low density situations with few interacting pedestrians to train the algorithms. A collection of datasets that are widely used to train and evaluate the algorithms are TrajNet++ \cite{kothari2021human} or OpenTraj \cite{amirian2020opentraj}. The KB models do not need data to learn pedestrian behavior. Data is just needed for calibrating the parameters. In practice, this is often done by using experimental data including high density situations, see, e.g., \cite{ExpJuelich}. The data used for the KB approach contains the trajectories of each pedestrian as well as the structure of the environment and locations of obstacles. For the DL algorithms, the data can be more varied because the algorithms are more flexible in terms of inputs. In the graph neural networks, the pedestrians and their interactions are described through a graph \cite{vemula2018social}. Other authors use spatial information represented as points of interest \cite{bartoli2018context} or as occupancy maps \cite{Pfeiffer2018}. By using CNNs it is even possible to use images or videos as inputs for the predictions \cite{sophie, kosaraju2019}.
Without sufficient amounts of data, the DL algorithms are not capable to learn pedestrian behavior and predict future trajectories successfully. In addition to the amount of data, pre-processing is an important step to archive good results and efficiently train the algorithm. One pre-processing technique is data normalization where the coordinates of the trajectories are normalized to coordinates with origin in the first observation, coordinates with origin in the last observation, or relative coordinates \cite{zamboni2022pedestrian}. Another important pre-processing technique is data augmentation. Using GANs \cite{shorten2019survey} is one form of data augmentation, but there are also more basic forms like rotating the input trajectories, mirror the trajectories, or applying Gaussian filter \cite{zamboni2022pedestrian}. Schöller et al. \cite{scholler2020constant} show that data augmentation helps to prevent the DL algorithm from learning environmental priors instead of pedestrian behavior. 

\paragraph{Numerical comparison}
Many studies report on numerical comparisons of KB and DL approaches for the prediction of trajectories. 
Authors generally used average and final displacement errors (ADE and FDE) to quantify the prediction accuracy. 
Several articles include the social-LSTM \cite{Alahi2016}  and the social force model \cite{Helbing1995} as references for, respectively, DL and KB approach.
In Table~\ref{tab:7}, the ADE and FDE metrics from selected studies using the social-LSTM as a benchmark are shown. 
All these articles use the same algorithm, the same datasets, the same error metrics, the same length of observed (3.2s) and predicted trajectories (4.8s), and the same cross-validation methods. 
Nevertheless, the results vary significantly from one analysis to another. 
This is not just a problem of the social-LSTM, but of the evaluation of the DL approach in general. 
Reasons for this high variation could be the randomness of the dataset splitting, the randomness of the starting coefficients, differences in the implementation of the algorithms, or different hyperparameter settings for the training, among others. 
An attempt to solve this problem is the trajectory forecasting challenge Trajnet++ \cite{kothari2021human}. 
The challenge provides a uniform sampling and evaluation system allowing to compare rigorously different algorithms in the same framework. 
Table~\ref{tab:8} reports on ADE and FDE metrics for selected articles using the social force model and social-LSTM algorithm.
We can observe that the error estimates also vary significantly with the social force model.
Yet, in contrast to Table~\ref{tab:7}, the studies do not systematically use the same dataset and prediction length making direct comparisons potentially biased. 
The objective is to compare the KB and DL approaches using relative errors. 
Assuming that the setting for KB and DL approaches are identical for a given study, the comparison of relative errors is fair.
The variations from one study to another are again very significant.
However, except in the work of Cheng et al. \cite{cheng2021amenet}, the social-LSTM systematically outperforms the social force model in terms of prediction accuracy. 
The mean ADE and FDE relative errors are, respectively, 105 and 80.4\%.
In the work from Hasan et al. \cite{hasan2018mx} or from Song et al. \cite{song2020pedestrian} the differences are especially huge, the social-LSTM being up to 341\% more accurate. 
Such results are statistically not surprising since the DL approach is based on much more free parameters (coefficients) than the KB approach.

\begin{table*}[!ht]
    \caption{Quantitative comparison of ADE and FDE metrics for articles using social-LSTM as benchmark with different datasets.}
    \centering
    \begin{tabular}{l|c|c|c|c|c|c}
    \hline
    \textbf{First author, year}&\textbf{Average}&\textbf{ETH}&\textbf{HOTEL}&\textbf{ZARA1}&\textbf{ZARA2}&\textbf{UCY}\\
    \hline
    Syed, 2019 \cite{syed2019sseg}&0.08 / 0.14&0.15 / 0.295&0.05 / 0.08&0.05 / 0.08&0.07 / 0.1&0.1 / 0.16\\
    Xue, 2018 \cite{xue2018ss}&0.12 / 0.17&0.2 / 0.37&0.08 / 0.13&0.08 / 0.11&0.07 / 0.12&0.2 / 0.24\\

    Manh, 2018 \cite{manh2018scene}&0.25 / 0.22&0.18 / 0.34&0.25 / 0.29&0.37 / 0.33&0.19 / 0.1&0.25 / 0.03\\
    Alahi, 2016 \cite{Alahi2016}&0.27 / 0.61&0.5 / 1.07&0.11 / 0.23&0.22 / 0.48&0.25 / 0.5&0.27 / 0.77\\
    Vemula, 2018 \cite{vemula2018social}&0.37 / 3.32&0.46 / 4.56&0.42 / 3.57&0.21 / 0.65&0.41 / 3.39&0.36 / 4.45\\
    Hossain, 2022 \cite{hossain2022sfmgnet}&0.44 / 0.98&0.60 / 1.31&0.15 / 0.33&0.43 / 0.93&0.51 / 1.09&0.52 / 1.25\\
    Zhu, 2019 \cite{zhu2019starnet}&0.45 / 0.91&0.73 / 1.48&0.49 / 1.01&0.27 / 0.56&0.33 / 0.7&0.41 / 0.84\\
    Hasan, 2018 \cite{hasan2018mx}&0.64 / 1.45&---&---&0.68 / 1.53&0.63 / 1.43&0.62 / 1.40\\
    Gupta, 2018 \cite{gupta2018}&0.72 / 1.54&1.09 / 2.35&0.79 / 1.76&0.47 / 1.00&0.56 / 1.17&0.67 / 1.40\\
    \end{tabular}
    
    \label{tab:7}
\end{table*}

\begin{table*}[!ht]
    \caption{Quantitative comparison of the social force model and the social-LSTM (average ADE/average FDE).}
    \centering
    \begin{tabular}{l|c|c|c|c|c|c}
    \hline
    \multicolumn{1}{r|}{\textbf{First author}}&Alahi \cite{Alahi2016}&Fernando \cite{fernando2018soft} &Cheng \cite{cheng2021amenet}&Fernando \cite{fernando2018gd} &Hasan \cite{hasan2018mx}&Song \cite{song2020pedestrian} \\[-2mm]
    \textbf{Approach}&&&&&&\\
    \hline
    Social force model$\qquad\quad$&0.39 / 0.60&3.24 / 4.86&0.37 / 1.27&1.5 / 2.46&4.28 / 7.63&0.61 / 0.96\\
    Social-LSTM&0.27 / 0.61&1.76 / 3.51&0.67 / 3.1&1.1 / 1.89&0.97 / 2.08&0.25 / 0.22\\
    \hline
    Relative error [\%]&44 / -1.6&84 / 38&-80 / -145&36 / 24&341 / 267&144 / 336\\
   
    \end{tabular}
    
    \label{tab:8}
\end{table*}

\paragraph{Advantages and disadvantages of KB and DL approaches}

The last paragraph shows that the DL algorithms outperform the KB models in terms of prediction accuracy of pedestrian trajectories in low density situations and that the results of the error metrics have a high fluctuation.
This fluctuation is mainly a problem of the DL algorithms because the evaluation of these algorithms is based on these error metrics. In addition to this lack of reproducibility, there is also the problem of missing explainability of the DL approach. This means that the coefficients can not be physically understood and interpreted. Therefore it is not clear why the predicted trajectories have the given shapes. In applications where autonomous systems make decisions, it is important to understand and communicate why specific choices have been made \cite{pasquale2017toward}. But these in large numbers existing coefficients are also an advantage of the DL approach because they make it possible to learn complex behaviors by fitting a set of values such that the predicted behavior fits the observed behavior. These algorithms do not need a priori knowledge about the system, just observations.

The comparison of the KB models with the DL algorithms is mostly done numerically, disregarding the missing reproducibility and high fluctuation. Furthermore, important advantages of the KB models are not taken to account in the numerical comparison. First of all, the KB models have the advantage of simple forms and interpretable parameters, which makes it easy to reproduce the results and to understand the predictions. The interpretability of the parameters makes the models flexible and easy to cope with environmental and behavioral changes. For example, if the preferences of pedestrians change because of a pandemic, which entails social distance and desired time-gap to rise, the KB methods can change the parameters to model these new situations, without needing new data. In the DL approach, new data is necessary to learn behavioral changes. But the KB models also have disadvantages, like the need of domain knowledge. Furthermore, only average pedestrian behavior is simulated and it is difficult to capture the complete crowd behavior range with a single model \cite{moussaid2011simple}.
The advantages and disadvantages, as well as other differences regarding the applications of both KB and DL approaches, are summarized in Table~\ref{tab:9}.

\begin{table*}[!ht]
      \caption{Application oriented comparison of the KB and the DL approach for prediction of pedestrian dynamics.}
  \label{tab:9}%
  \centering
    \begin{tabular}{p{9em}p{17em}p{17em}}
    \hline
        & \textbf{Knowledge-based} & \textbf{Deep learning}\\
          \hline
          \emph{Applications} & Simulation tool & Socially-aware mobile robots \\
          & Infrastructure design&Design of intelligent tracking systems\\
          & Evacuation situations & Pedestrian trajectory prediction for autonomous vehicles\\[-3mm]
          &Intelligent transport systems&\\
   \hline
   \emph{Application scales}&Large-scale simulation&Local prediction scale (in time and space)\\
   &Large infrastructures (train station, stadium, commercial mall), urban centers, outdoor events&Few interacting pedestrians\\

      \hline
           \emph{Crowd density} &Low density (long range interaction, e.g., collision avoidance models)  &Low density situations\\[-3mm] 
           &&Long range pedestrian interaction\\[-1mm]
           &High density (short range interaction, e.g., force-based models)& \\
   \hline
    \emph{Advantages} & Interpretable parameter & Accurate predictions\\
          & Explainable predictions& Learn complex interaction\\
          & Reproducible& No domain knowledge necessary\\
          &Few data needed & No modelling-bias\\
          & & Can process different types of data \\
         
    \hline
    \emph{Disadvantages} & Low use of data & Not interpretable\\
          & Averaged behavior only& Not reproducible in praxis\\
          & Not suitable to complex interaction&Lack of generalisation\\
          & Complete crowd behavior range with a single model difficult to capture&Require large amount of data\\[-3mm]
          &&Necessary complexity of the network unknown\\

        \hline
    \end{tabular}%
\end{table*}%

\newpage
\section{Future directions}
\label{sec:5}
In the last part of this work we want to look into the future and describe which trends we can identify.

\subsection{The hybrid approach}
\label{sec:5.1}
In Table~\ref{tab:8} it is shown that many points that can be criticized in the predictions with the DL algorithms are strengths of the KB models. KB models have few parameters with physical interpretations requiring few data for calibration, while the coefficients of the DL algorithms are generally not interpretable and much data is needed in the training phase.
For this reason, the combination of both approaches seems to have potential as some pioneer studies point out \cite{pedreschi2019meaningful,hu2016harnessing}. This combination is called the hybrid approach and even though the idea of combining both approaches has picked up momentum just in the last few years, there is already a vast amount of work in diverse disciplines. Examples of the hybrid approach are given in applications like the discovering of novel climate patterns \cite{kawale2013graph,faghmous2015daily}, the finding of novel compounds in material science \cite{hautier2010finding,fischer2006predicting}, the designing density functionals in quantum chemistry \cite{li2016understanding}, or the improving imaging technologies in bio-medical science \cite{wong2009active,xu2015robust}. Recently, the first applications for pedestrian trajectory predictions can be found in \cite{8950093,antonucci2020generating, ALAHi2021anchors}.

In general, there are three ways of combining both KB and DL approaches. First, KB models can be used to improve the DL algorithms.
One possibility to do so is \emph{data generation}. Simulations based on KB models are carried out to obtain a synthetic data set, that is used to train and test the neural network. A popular application of data generation can be found in the training of autonomous vehicles for the augmentation of data for scenarios that are not sufficiently represented in the available data set \cite{lee2018spigan, von2020combining}. In \cite{9053301,8950093,Alahi2016}, the authors apply the social force model to generate a synthetic data-set for the training of neural networks and setting of hyper-parameters.
\emph{Knowledge-guided design of architecture} is another possibility to improve the DL algorithms with knowledge. The modular and flexible nature of the networks enables the use of knowledge to specify node connections that capture dependencies among variables. In this way, Antonucci et al. \cite{antonucci2020generating} successfully embed the social force model in the architecture of a neural network to generate predictions of human motion.
A common technique to improve the output of DL algorithms is \emph{knowledge-guided loss function}. It makes the output consistent with physical laws so that unrealistic prediction can be ruled out \cite{willard2020integrating,silvestri2021injecting}. Because the training of DL algorithms is an iterative process, they require an initial choice of coefficients as a first step to commence the learning process. \emph{Knowledge-guided initialization of the network} can be used to guide the network at an early stage to archive generalizable and physically consistent results \cite{karpatne2017theory}.

The second way of combining the approaches is obtained by using DL algorithms to improve the prediction accuracy of the KB models.
One of the oldest and most common way for addressing the imperfections of the KB approach is \emph{residual modeling}. A DL algorithm learns to correct the errors made by the KB model by predictions the model residuals \cite{willard2020integrating}. Bahari et al. \cite{BAHARI2021103010} proposed a so-called "realistic residual block" to improve vehicle trajectory predictions. Another way to improve the KB predictions consists in calibrating the parameters of the KB models by using DL algorithms. Göttlich et al. \cite{gottlich2020artificial} use neural networks to estimate the parameters of the social force model. Hossain et al. \cite{hossain2022sfmgnet} extent the social force model \cite{Helbing1995} using group forces, based on neural networks, to consider interactions with static obstacles, other pedestrian, and pedestrian groups. Kreiss \cite{kreiss2021deep} uses deep neural networks to estimate different interaction potentials for the social force model.

Given enough data, DL algorithms are capable of predicting pedestrian trajectories of a given scene with relative high accuracy. However, pedestrians experience different interactions in different situations. Whether they are able to make accurate predictions for the hole range of possible scenes and interactions is a open question. The hybrid approach is useful for this problem because it can compensate scarcity of data with available knowledge. The hybrid approaches are promising to improve the predictions of pedestrian trajectories because they benefit from the strengths of both approaches and reduce shortcomings like the missing explainability of the DL approach.

\subsection{Other directions}
\label{sec:5.2}
As it has been shown in Fig.~\ref{fig:2}, the DL approach has gained attention just a few years ago. At the current state-of-the-art, DL is mostly used for learning human behavior and predicting single pedestrian trajectories. 
In the future, the DL approach could be used in more applications in the discipline of pedestrian dynamics and notably for large scale simulation and the simulation of collective dynamics. 
Nowadays, the behavior and interactions of agents in simulation platforms currently available rely on KB approaches \cite{Korhonen2010,Curtis2016,Taillandier2019}. 
Following the success of the DL algorithms for predictions of pedestrians in low density situations, there is a high potential to use these methods for accurate predictions of crowd dynamics like in evacuation situations. 

In addition to the KB models, RL algorithms are often used for simulating crowd dynamics, but they have some disadvantages, like the need of a reward function, a priori given goal/destination, or the difficulty to incorporate interactions. 
Such difficulties are overcome by the supervised DL algorithms. The combination of RL and DL algorithms seems to be promising to solve these disadvantages as Everett et al. \cite{everett2021collision}  have recently shown. Other works have shown successful applications of deep RL for crowd simulations \cite{lee2018crowd} or even evacuation dynamics \cite{zhang2021crowd}.

There are still open questions to be tackled, before supervised DL algorithms could be successfully used for large-scale crowd simulations, like: 
\begin{itemize}
    \item Which kind of neural networks, which complexity and which training data should be used for large-scale simulation including different types of geometries?
    \item Should the type and complexity of appropriate networks as well as the data used for the training depend on the scenarios of the simulation?
\end{itemize}
Besides these considerations, one may expect to develop deep networks trained on large amounts of data that could predict accurate trajectories for any density levels and any type of facility. 
Such networks could be used in agent-based simulation platforms whose objective is to simulate any type of scenario.
Yet, the question whether such universal networks, as supervised approaches, require training on datasets representing the full diversity of pedestrian dynamics remains open.
It may be possible, as in unsupervised approaches, that training on few data is sufficient to obtain accurate predictions even for scenarios the networks are not trained for. 
Such question raises more generally on the robustness of the predictions against new data, i.e., new situations, scenarios, density levels, or types of facilities. 
Some preliminary results obtained in a corridor and a bottleneck have shown that neural networks are quite robust to new types of facilities and may even overcome KB models in terms of prediction robustness \cite{tordeux2017b,tordeux2020}.

Another possible development direction of DL approaches for prediction pedestrian trajectory concerns the nature and type of the inputs. 
The inputs of the algorithms are mostly the trajectories of the pedestrian over finite past horizon.
Yet, studies of pedestrian dynamics with KB models identified different relevant variables like, besides the relative position \cite{Helbing1995}, the relative velocity \cite{Chraibi2010}, the time gap \cite{moussaid2011simple}, or indicators related to possibilities of collision such as bearing angle \cite{Ondrej2010}, collision cone \cite{vanderBerg2008} or time-to-collision \cite{Karamouzas2014}.
The use of these variables as inputs, even if they can theoretically be deduced from the trajectories, could allow obtaining prediction improvement, especially in the case of low amount of data for the training phase. 
We can already observe a trend to use more inputs and notably the speed difference \cite{Ma2016,tordeux2020,kothari2021human} and other hidden variables estimated by training \cite{gupta2018,socialway}. 
One may expect to include further variables as inputs. 
The time-to-collision and the time gap, whose roles are fundamental in KB pedestrian dynamic models \cite{Karamouzas2014,moussaid2011simple}, are promising candidates. 
Such variables contain information about the relevance of the interactions and may even substitute the interaction or attention modules of DL approaches.

\section*{Acknowledgment}
  The authors acknowledge the Franco-German research project MADRAS funded in France by the Agence Nationale de la Recherche (ANR, French National Research Agency), grant number ANR-20-CE92-0033, and in Germany by the Deutsche Forschungsgemeinschaft (DFG, German Research Foundation), grant number 446168800. 


\ifCLASSOPTIONcaptionsoff
  \newpage
\fi


\end{document}